\documentclass[sigconf,natbib=true,anonymous=false]{acmart}

\usepackage{graphicx}
\usepackage{array}
\usepackage{makecell}
\usepackage{multirow}
\usepackage{booktabs}
\usepackage{amsmath}
\usepackage{dblfloatfix}
\usepackage{soul}
\usepackage{subfig}

\DeclareMathOperator*{\argmin}{argmin}

\AtBeginDocument{%
  \providecommand\BibTeX{{%
    \normalfont B\kern-0.5em{\scshape i\kern-0.25em b}\kern-0.8em\TeX}}}

\setcopyright{iw3c2w3}
\copyrightyear{2023}
\acmYear{2023}
\acmDOI{XXXXXXX.XXXXXXX}

\acmConference[Preprint]{Arxiv}{2023}{Arxiv}
\acmPrice{15.00}
\acmISBN{978-1-4503-XXXX-X/18/06}

\begin{document}

\title{iQPP: A Benchmark for Image Query Performance Prediction}

\author{Eduard Poesina}
\affiliation{%
  \institution{Department of Computer Science, University of Bucharest}
  \streetaddress{14 Academiei}
  \city{Bucharest}
  \country{Romania}
  \postcode{010014}
}
\email{eduardgabriel.poe@gmail.com}

\author{Radu Tudor Ionescu}
\authornote{Corresponding author.}
\affiliation{%
  \institution{Department of Computer Science, University of Bucharest}
  \streetaddress{14 Academiei}
  \city{Bucharest}
  \country{Romania}
  \postcode{010014}
}
\email{raducu.ionescu@gmail.com}

\author{Josiane Mothe}
\affiliation{%
  \institution{INSPE, IRIT UMR5505 CNRS, Universit\'{e} Toulouse Jean-Jaur\`{e}s}
  \streetaddress{118 Rte de Narbonne}
  \city{Toulouse}
  \country{France}
  \postcode{F-31400}
}
\email{josiane.mothe@irit.fr}

\renewcommand{\shortauthors}{Poesina et al.}

\begin{abstract}
To date, \emph{query performance prediction} (QPP) in the context of {content-based image retrieval}  remains a largely unexplored task, especially in the {query-by-example} scenario, where the query is an image. To boost the exploration of the QPP task in image retrieval, we propose the first benchmark for \emph{image query performance prediction} (iQPP). First, we establish a set of four data sets (PASCAL VOC 2012, Caltech-101, ROxford5k and RParis6k) and estimate the ground-truth difficulty of each query as the average precision or the precision@$k$, using two state-of-the-art image retrieval models. Next, we propose and evaluate novel pre-retrieval and post-retrieval query performance predictors, comparing them with existing or adapted (from text to image) predictors. The empirical results show that most predictors do not generalize across evaluation scenarios. Our comprehensive experiments indicate that iQPP is a challenging benchmark, revealing an important research gap that needs to be addressed in future work. 
We release our code and data as open source at \url{https://github.com/Eduard6421/iQPP}, to foster future research.
\end{abstract}

\ccsdesc[500]{Information systems~Information retrieval query processing}
\ccsdesc[500]{Information systems~Information retrieval}
\ccsdesc[500]{Information systems~Test collections}
\ccsdesc[500]{Information systems~Retrieval effectiveness}
\ccsdesc[300]{Information systems~Retrieval tasks and goals}

\maketitle

\section{Introduction}

Query performance prediction (QPP) (also known as query difficulty prediction or query difficulty estimation) is the task of estimating the effectiveness of a set of search results retrieved in response to a query, without relevance judgments \cite{CronenTownsend-SIGIR-2002}. QPP is extremely important in identifying poorly performing queries, which might require the use of a more powerful retrieval system or a query expansion method to improve the search results. By predicting the performance of queries, the system can optimize the search, presenting the most relevant information to the user. This can lead to a better user experience and increased satisfaction. Furthermore, a better understanding of query difficulty can also suggest improvements to the underlying algorithms and systems used in information retrieval, leading to more effective and efficient systems in the future. Hence, QPP is a crucial component for achieving effective retrieval results and improving the overall search experience. The importance of QPP has been widely recognized in text retrieval \cite{CronenTownsend-SIGIR-2002,He-SPIRE-2004,Mothe-SIGIR-2005,Hauff-CIKM-2008,Hauff-ECIR-2009,Shtok-SIGIR-2010,Cummins-SIGIR-2011,Kurland-CIKM-2012a,Kurland-CIKM-2012b,Shtok-TIS-2012,Cummins-TIS-2014,Katz-SIGIR-2014,Raiber-SIGIR-2014,Tao-CIKM-2014,Shtok-TIS-2016,Roitman-SIGIR-2017a,Roitman-SIGIR-2017b,Chifu-SIGIR-2018,Mizzaro-SIGIR-2018,Roitman-SIGIR-2018,Zamani-SIGIR-2018,Roy-IPM-2019,Zendel-SIGIR-2019,Arabzadeh-IPM-2020,Dejean-SAC-2020}, being an actively studied task nowadays \cite{Arabzadeh-CIKM-2021,Arabzadeh-ECIR-2021,Chen-ECIR-2022,Datta-SIGIR-2022,Datta-WSDM-2022,Datta-TIS-2022,Faggioli-IRJ-2022,Jafarzadeh-IPM-2022,Sarwar-ADCS-2021}. However, in the context of content-based image retrieval (CBIR), the QPP task received comparably lower attention from the research community, with only a few studies addressing the topic \cite{Xing-ECIR-2010,Li-NC-2012,Nie-TIS-2012,Tian-TMM-2012,Jia-ICME-2014,Jia-SP-2015,Tian-TMM-2015,Pedronette-SIBGRAPI-2015,Sun-TIP-2018,Valem-ICMR-2021}. Furthermore, only a handful of papers \cite{Li-NC-2012,Pedronette-SIBGRAPI-2015,Sun-TIP-2018,Valem-ICMR-2021} study query difficulty prediction in the query-by-example scenario, where the query is an image.

Since we consider that QPP in text and image retrieval is equally important, the goal of this work is to raise the level of exploration of QPP in the image domain to the same level of exploration currently observed in the text domain. To this end, we propose the first benchmark for image query difficulty prediction, which we term iQPP, in the context of query-by-example content-based image retrieval, where images have to be retrieved given an image query. Our benchmark comprises four data sets (PASCAL VOC 2012 \cite{Everingham-IJCV-2015}, Caltech-101 \cite{Li-CT-2022}, ROxford5k \cite{Radenovic-CVPR-2018} and RParis6k \cite{Radenovic-CVPR-2018}), two image retrieval systems~\cite{Radenovic-TPAMI-2019,Revaud-ICCV-2019}, as well as several pre-retrieval and post-retrieval query performance predictors, for which we deliver the predicted and ground-truth performance levels for two effectiveness measures. The data sets are chosen based on their popularity, aiming to accommodate a high variety of images, from landmark photos to pictures of various object classes. The retrieval systems are chosen due to their state-of-the-art performance coupled with the availability of the open source models. Since research on image QPP is scarce, we turn our attention to proposing several novel predictors along with our benchmark. First of all, we propose four novel pre-retrieval predictors, namely $(i)$ the magnitude of the reconstruction error of denoising \cite{Vincent-JMLR-2010} or masked \cite{He-CVPR-2022} auto-encoders trained on the database, $(ii)$ the density of the k-means cluster to which the query image embedding is assigned, $(iii)$ the confidence distribution of a classification head attached to the embedding layer of the retrieval model, and $(iv)$ the score predicted by a fine-tuned ViT model \cite{Dosovitskiy-ICLR-2020}. We note that the first three pre-retrieval predictors are unsupervised, while the last one is supervised on labeled training queries. Second of all, we propose four novel post-retrieval predictors, namely $(v)$ a query feedback method redesigned for the image domain, $(vi)$ the intersection over union (IoU) ratio for the results retrieved while iteratively removing the most discriminative features, $(vii)$ the dispersion (variance) of the embeddings of the retrieved results, and $(viii)$ the difficulty score predicted by a regression model applied on all the other predictors. Among the proposed post-retrieval predictors, only the last one is supervised. We compare the proposed predictors with existing methods for image difficulty \cite{Ionescu-CVPR-2016,Soviany-CVIU-2021} or query performance \cite{Sun-TIP-2018} prediction, as well as the standard deviation of predicted relevance scores, which was originally proposed for QPP in the text domain \cite{Cummins-SIGIR-2011}.

We carry out a comprehensive set of experiments on the four aforementioned data sets to evaluate the capacity of the designated predictors in predicting the ground-truth difficulty of the queries. As in previous studies~\cite{Hauff-CIKM-2008,Datta-SIGIR-2022}, we employ the Pearson correlation and the Kendall $\tau$ correlation as evaluation measures. Our results indicate that supervised and post-retrieval predictors tend to achieve better performance. However, the empirical evidence shows that the image QPP task is far from being solved, since none of the existing or proposed predictors can surpass a Kendall $\tau$ of $0.65$, and there is no predictor to consistently surpass its competitors on all four data sets. This observation indicates that there is an important research gap in image QPP, which needs to be addressed in future work. To ensure the reproducibility of the results and foster future research, we release our code and data as open source at \url{https://github.com/Eduard6421/iQPP}.

In summary, our contributions are threefold:
\begin{itemize}
    \item We develop the first benchmark for image query performance prediction in the query-by-example CBIR setting.
    \item We propose eight novel pre-retrieval and post-retrieval image query performance predictors.
    \item We present extensive experiments on the four image data sets included in our benchmark.
\end{itemize}
 
\section{Related Work}

\subsection{QPP in ad-hoc text retrieval}

Query performance prediction became popular in the mid-2000s in ad-hoc text retrieval~\cite{CronenTownsend-SIGIR-2002,Carmel-SIGIR-2006}. Since then, researchers explored QPP using a broad range of approaches, giving rise to various categories of predictors. For example, query performance predictors can be categorized into pre-retrieval and post-retrieval. Pre-retrieval predictors aim to predict query performance prior to querying the document collection, while post-retrieval ones imply carrying out a search before estimating effectiveness~\cite{Carmel-SIGIR-2006}. Some popular pre-retrieval predictors are term rareness, specificity or distribution~\cite{DeLoupy-LREC-2000,Scholer-JASIST-2004,He-IS-2006,Zhao-ECIR-2008}, term ambiguity~\cite{DeLoupy-LREC-2000,Mothe-SIGIR-2005} and query complexity~\cite{Mothe-SIGIR-2005,Goeuriot-SIGIR-2014}. Hauff et al.~\cite{Hauff-CIKM-2008} analyzed 22 pre-retrieval predictors 
and concluded that the predictors jointly considering the query and the collection are better than the ones only  considering the query. They also observed that the performance of predictors depends on the test collection and the underlying retrieval model. 
Due to the extra available information, post-retrieval predictors are usually more effective, although less efficient~\cite{Shtok-TIS-2012}. Many of these predictors mainly consider the homogeneity, magnitude or variance of the retrieved document scores \cite{CronenTownsend-SIGIR-2002,Sehgal-WPQDMA-2005,Carmel-SIGIR-2006,Shtok-TIS-2012,Zhang-APWeb-WAIM-2018}. For example, the clarity score~\cite{CronenTownsend-SIGIR-2002} measures the term distribution probability of the retrieved documents and the term distribution probability of the whole collection~\cite{CronenTownsend-SIGIR-2002}. Other post-retrieval predictors measure the divergence or stability of the retrieved document list when the query is perturbed using relevance feedback~\cite{Zhou-SIGIR-2007}, sub-queries \cite{YomTov-SIGIR-2005} or different scoring functions~\cite{Aslam-ECIR-2007}. 
Individual predictors may have high variance~\cite{Chifu-SIGIR-2018,Zamani-SIGIR-2018,Jafarzadeh-IPM-2022} and are not robust across collections. Thus, some studies have  considered combining predictors. For non-supervised methods, studies have focused on analyzing classes of queries or the link between system/query features and  effectiveness, using for example factorial analysis \cite{Dincer-JASIST-2007,Bigot-IR-2011}. Supervised methods represent however the most common means to combine predictors in QPP, comprising approaches based on linear combinations~\cite{Zhou-CIKM-2006,Chifu-SIGIR-2018,Roy-IPM-2019,Jafarzadeh-IPM-2022}, genetic algorithms~\cite{Bashir-AI-2014}, and neural networks \cite{Zamani-SIGIR-2018,Datta-WSDM-2022}. Supervised predictors are usually evaluated using k-fold or leave-one-out cross-validation~\cite{Carmel-SIGIR-2006}.




To our knowledge, there is no clear benchmark for text QPP. However, there are some common practices. First, several collections are used in a study, allowing researchers to evaluate the robustness of a predictor across collections. QPP for ad-hoc search relies mostly on TREC collections\footnote{Text REtrieval Conference (\url{http://trec.nist.gov})} \cite{Chifu-SIGIR-2018,Hauff-CIKM-2008,YomTov-SIGIR-2005}. Evaluation encompasses several aspects, such as the measure used to assess the retrieval system performance, and the one used to evaluate the QPP accuracy.
As performance measures, researchers employ the usual ad-hoc retrieval effectiveness measures, e.g.~average precision (AP), NDCG, or precision at a certain cut-off point of the retrieved list \cite{Chifu-SIGIR-2018,Hauff-CIKM-2008,Jafarzadeh-IPM-2022,Roitman-SIGIR-2019,YomTov-SIGIR-2005,Zhao-ECIR-2008}. The prediction accuracy is measured by considering the actual effectiveness and the predicted performance. To evaluate this relation, most of the studies consider the Pearson correlation. Since the link between the two measures may not be linear, Kendall and Spearman correlations are often employed as additional measures \cite{Chifu-SIGIR-2018,Hauff-CIKM-2008,Jafarzadeh-IPM-2022,Roitman-SIGIR-2019,YomTov-SIGIR-2005,Zhao-ECIR-2008}. 
We follow similar evaluation principles in constructing our iQPP benchmark. 

Different for the mainstream area studying QPP in ad-hoc text retrieval, we underline that studies on QPP in image retrieval require new predictors adapted to the image domain, as well as distinct data sets, containing images instead of text documents. Hence, research on image QPP naturally diverges to a different direction, specific to the image domain. We cover this direction in the next section.

\subsection{QPP in image search}


One of the first contributions on QPP in the image domain is the work of Xing et al.~\cite{Xing-ECIR-2010}. The authors used query words and context information to compute a set of four text-based pre-retrieval features and  train a model for QPP in image retrieval.
Subsequent studies turned their attention to post-retrieval predictors.
Tian et al.~\cite{Tian-TMM-2012} focused on QPP for web image search, where the query is a piece of text and the results are images. 
The authors proposed the visual clarity score inspired from the clarity score defined for texts~\cite{CronenTownsend-SIGIR-2002}, which measures the difference in the distribution of the top retrieved images and the whole collection. They also use the coherence score based on the visual similarity among the retrieved images. In a later study, Tian et al.~\cite{Tian-TMM-2015} introduced an approach to reconstruct an image query based on the images returned in response to a text query. They estimated query performance via the differences between the ranked lists of the text query and the reconstructed image query. Nie et al.~\cite{Nie-TIS-2012} presented a two-stage pipeline for QPP. In the first stage, ranked image lists are classified into person-related and non-person-related. In the second stage, the relevance probability of the query is estimated via graph-based learning, as well as visual content. The authors adapt the visual content representation to the class predicted in the first stage. Jia et al.~\cite{Jia-ICME-2014,Jia-SP-2015} introduced a post-retrieval predictor that divides the retrieved images into pseudo-positive and pseudo-negative via pseudo-relevance feedback. Next, a voting scheme is applied to label the images as relevant or not. The pseudo-relevance labels are further used to provide an estimate for the AP.


To our knowledge, there are only a few studies \cite{Li-NC-2012,Pedronette-SIBGRAPI-2015,Sun-TIP-2018,Valem-ICMR-2021} that try to predict the performance of image queries. Li et al.~\cite{Li-NC-2012} proposed a post-retrieval predictor that examines the top ranked images using the clarity score, the spatial consistency of local descriptors, and the appearance consistency of global features. The method is specifically designed for image retrieval models based on the bag-of-visual-words \cite{Ionescu-KTCVTM-2016,Philbin-CVPR-2007}. 
Pedronette et al.~\cite{Pedronette-SIBGRAPI-2015} proposed an unsupervised post-retrieval predictor based on the cluster hypothesis \cite{Kurland-ECIR-2014}, considering that the images belonging to a highly effective ranked list should appear in the ranked lists of each other. The authors study different ways to measure the density of reciprocal references among retrieved images, conducting experiments on relatively small data sets (each of about 1,000 images). Sun et al.~\cite{Sun-TIP-2018} proposed a supervised post-retrieval predictor that transforms the ranked list of images into a similarity or correlation matrix which is further given as input to a convolutional neural network (CNN). Valem et al.~\cite{Valem-ICMR-2021} extended the work of Sun et al.~\cite{Sun-TIP-2018} by generating synthetic ranked lists as training data for the CNN, requiring a more complex training procedure.

As related studies on QPP in ad-hoc text retrieval, methods studying QPP in image search commonly employ AP and NDCG as effectiveness measures, and the Pearson coefficient as query performance prediction measure \cite{Li-NC-2012,Nie-TIS-2012,Pedronette-SIBGRAPI-2015,Sun-TIP-2018,Valem-ICMR-2021}.


Most of the above studies, e.g.~\cite{Xing-ECIR-2010,Nie-TIS-2012,Tian-TMM-2012,Jia-ICME-2014,Jia-SP-2015,Tian-TMM-2015}, predict the performance of text queries in CBIR.
Our study is among the few works \cite{Li-NC-2012,Pedronette-SIBGRAPI-2015,Sun-TIP-2018,Valem-ICMR-2021} studying the performance of image queries. To the best of our knowledge, we are the first to propose a benchmark for image QPP in image retrieval. The benchmark includes four data sets, two retrieval models, and twelve predictors. Among the considered predictors, there are eight novel QPP approaches, adding to the novelty of our study. Furthermore, we are the first to propose pre-retrieval predictors for image queries.

\section{Problem Formulation}
\label{sec_formulation}

Let $\mathcal{D}=\{x_1, x_2, ...., x_n\}$ be a collection (database) of images, where $n$ is the number of images inside the collection. Given a query image $q$, an image retrieval system $R$ returns a ranked list of $k$ images denoted as $\rho_{q,R}=\left[x^{(q,R)}_1,x^{(q,R)}_2,...,x^{(q,R)}_k\right]$, where $k \leq n$ and $x^{(q,R)}_i \in \mathcal{D}$ is the $i$-th most similar image to $q$, as evaluated by $R$. A retrieval model is a tuple $R=(f,\delta)$, where $f\!:\mathbb{R}^{h\times w}\rightarrow \mathbb{R}^d$ is a model, e.g.~neural network, that maps each input image $x_i$ of $h\times w$ pixels to a $d$-dimensional embedding vector $v_i$, and $\delta: \mathbb{R}^d \times \mathbb{R}^d \rightarrow \mathbb{R}$ is a distance or similarity function. To obtain the ranked list, the model $f$ is applied to the query $q$ (online) and the images $x_i \in \mathcal{D}$ (offline). Then, the measure $\delta$ is applied on each pair of embedding vectors $v_q$ and $v_i$, and, based on the returned distance or similarity values, the images are sorted in descending or ascending order, respectively. 
Let $\rho_{q}=\left[x^{(q)}_1,x^{(q)}_2,...,x^{(q)}_k\right]$ denote the ground-truth ranked list of the most similar $k$ images from $\mathcal{D}$ to the query image $q$. Let $\mathcal{R}_k$ denote the set of all possible ranked lists of $k$ images. Let $P\!: \mathcal{R}_k \times \mathcal{R}_k \rightarrow \mathbb{R}$ be a performance measure, e.g.~average precision or precision@$k$, that estimates the effectiveness of the retrieval system $R$ on the query $q$, considering the ground-truth and predicted rankings. For any query image $q$ and retrieval model $R$, the goal of QPP is to predict the value returned by $P\left(\rho_{q}, \rho_{_{q,R}}\right)$, without having access to the ground-truth list $\rho_{q}$.

\section{Pre-Retrieval Predictors}
\label{sec:Pre}

To save space and avoid reiterating through the predictors presented here in a later section, we include hyperparameter choices in the current section. We use the same presentation format for the post-retrieval predictors described in Section~\ref{sec:Post}.

\subsection{Baselines}

The baselines presented below were previously used for generic image difficulty estimation \cite{Ionescu-CVPR-2016,Soviany-CVIU-2021}. We repurpose them as pre-retrieval query performance predictors.

\vspace{0.1cm}
\noindent
{\bf Generic image difficulty.} 
Ionescu et al.~\cite{Ionescu-CVPR-2016} proposed an image difficulty predictor trained on a data set collected with the help of human annotators. The ground-truth difficulty score of each image is based on measuring the average time taken by human annotators to search for different objects (present or missing) in the respective image. The image difficulty predictor is based on an ensemble of VGG networks \cite{Simonyan-ICLR-2014} pre-trained on ImageNet \cite{Russakovsky-IJCV-2015}, where the classification layer is replaced with a Support Vector Regression (SVR) model. We reproduce the original difficulty regressor of Ionescu et al.~\cite{Ionescu-CVPR-2016} and apply it on each query image. As discussed in \cite{Ionescu-CVPR-2016}, the image difficulty regressor scores images on a continuous scale, such that images with one object in a plain background receive low scores, and images with many objects in a complex scene (background) receive high scores.

\vspace{0.1cm}
\noindent
{\bf Number of objects divided by their area.}
In object detection, Soviany et al.~\cite{Soviany-CVIU-2021} observed that the number of detected objects and the total area covered by the objects are positively and negatively correlated with image difficulty~\cite{Ionescu-CVPR-2016}, respectively. In other words, an image depicting one object covering the entire image is easy for visual search, while an image depicting multiple objects covering a small area, e.g.~because the objects are photographed from far away, is difficult. Thus, following the intuition of Soviany et al.~\cite{Soviany-CVIU-2021} and transposing it to image QPP, the performance of an image query can be estimated as the number of objects divided by their average bounding box area. Let $\mathcal{B} = \{b_1, b_2, ..., b_m \}$ be the set of $m$ bounding boxes detected in query image $q$ by a pre-trained object detector, namely a Faster R-CNN \cite{Ren-NIPS-2015} with a ResNet-50 \cite{He-CVPR-2016} backbone. Let $h_i$ and $w_i$ denote the height and width of bounding box $b_i$. The difficulty score of an image $q$ is defined as follows:
\begin{equation}\label{eq_difficulty}
s(q,\mathcal{B}) = \frac{m}{\frac{1}{m}\sum_{i=1}^m w_i \cdot h_i}. 
\end{equation}

\subsection{Proposed}

\vspace{0.1cm}
\noindent
{\bf Auto-encoder reconstruction.} 
We train two types of auto-encoders (AEs) to reconstruct the images in the collection $\mathcal{D}$, namely denoising AEs \cite{Vincent-JMLR-2010} and masked AEs \cite{He-CVPR-2022}. A denoising AE corrupts input images with Gaussian noise and learns to reconstruct the original (uncorrupted) inputs. By adding noise, the AE avoids learning the identity mapping. Masked auto-encoders represent a modern attempt to learn discriminative representations by dividing input images into a grid of patches and eliminating a significant amount of patches (typically 75\%). A masked AE learns to reconstruct the missing patches. Both AE models embed images into a latent manifold that captures the most important patterns in the training data distribution. Auto-encoders are known for their capacity to represent images from the training distribution very well. However, as soon as an image from a different distribution is given as input, AEs exhibit poor reconstruction capabilities. We leverage this property and propose to train a denoising or masked AE on the collection of images $\mathcal{D}$ and apply it on the query image $q$. Intuitively, if the query image can be accurately reconstructed by the model, then the query is likely to be easy, i.e.~a low reconstruction error means that the query image belongs to the training data distribution. In contrast, a query image that is poorly reconstructed by the model indicates that it is not well represented by the training distribution. Moreover, we consider that the higher the reconstruction error, the farther away the query image is from the training distribution. 

The denoising AE is based on a convolutional architecture. The encoder is composed of four convolutional layers with ReLU activations and 2D batch normalization, while the decoder is formed of four convolutional layers with ReLU activations and nearest neighbor upsampling applied between layers. The masked AE is based on transformer blocks. The encoder uses an embedding dimension of $768$ and comprises $18$ transformer blocks, each with $16$ attention heads. The decoder is lighter, having only $8$ transformer blocks based on $512$-dimensional embeddings. We employ Adam \cite{Kingma-ICLR-1015} and the mean squared error (MSE) to train both AEs. We set the learning rate to $10^{-3}$ and the mini-batch size to $12$ for both models. During inference, the MSE is used as effectiveness score for the query images.

\vspace{0.1cm}
\noindent
{\bf K-means cluster density.}
We propose to cluster the embedding vectors given by the model $f$ for all images in the collection $\mathcal{D}$, using k-means clustering. The embedding of the query image $v_q=f(q)$ is assigned to one of the clusters denoted as $C_j$, represented by the centroid $c_j$, where $1 \leq j \leq K$ and $K$ is the number of clusters. We consider that the query is \emph{easy} if the cluster $C_j$ has many points densely packed together, and \emph{hard} if the cluster $C_j$ has only a few points spread over a large area. Aside from the cluster density, which is the same for all queries assigned to cluster $C_j$, the relation between a query and the cluster centroid provides another clue about the difficulty of the query. More precisely, the farther the embedding $v_q$ is from the cluster centroid, the more difficult the query. We combine the aforementioned conjectures into a closed form equation and compute the difficulty score $s(q)$ of query $q$ as follows:
\begin{equation}\label{eq_kmeans}
 s(q) = \frac{\delta(c_j, v_q) + var(C_j)}{|C_j|},   
\end{equation}
where $|C_j|$ is the cardinal of cluster $C_j$ and $var(C_j)$ is the variance of cluster $C_j$. We tune the hyperparameter $K$ taking values between 50 and 300, with a step of 50. We obtain optimal results with $K=150$. 


\vspace{0.1cm}
\noindent
{\bf Confidence distribution of self-supervised classification head.}
We propose to equip the embedding model $f$ with a softmax classification head and train it on the image database $\mathcal{D}$. After embedding each query with the model $f$, we additionally pass it through the classification head to obtain a class distribution. We conjecture that easy queries are likely to be assigned to one class with high confidence. At the same time, hard queries will be assigned to multiple classes with various confidence levels, indicating that the classifier is not certain what class label should be assigned to the query image. We alternatively employ two measures to estimate the class confidence distribution: dispersion and kurtosis. 

Since not all image collections have class labels attached, we train the classification head in a self-supervised manner. Following Caron et al.~\cite{Caron-ECCV-2018}, we first cluster the embedding vectors $v_i$ into $K$ clusters via k-means. Then, we use the cluster assignments as target class labels for our classification head. The head comprises two hidden layers of 50 neurons each with ReLU activations, and a softmax layer. We employ the Adam optimizer \cite{Kingma-ICLR-1015} with a learning rate of $10^{-4}$ to minimize the cross-entropy loss. During training, the embedding model $f$ is frozen. We tune the number of clusters in the range 50-300, at a step of 50. The optimal value for $K$ is 150.


\vspace{0.1cm}
\noindent
{\bf Fine-tuned ViT.}
Visual transformers \cite{Dosovitskiy-ICLR-2020} are a family of deep learning architectures that apply the self-attention mechanism to visual data. This allows the model to handle long-range dependencies and spatially-varying information in images, recently leading to improved performance in tasks such as image classification, segmentation, and generation. The power of these models relies on using a two-stage training process: $(i)$ large scale pre-training and $(ii)$ fine-tuning on downstream tasks. Following this procedure, we propose to fine-tune a visual transformer (ViT) model \cite{Dosovitskiy-ICLR-2020} to predict the performance levels of training query images. We select the ViT-B16 backbone pre-trained on ImageNet \cite{Russakovsky-IJCV-2015}, and fine-tune it on the QPP task for $100$ epochs with the Adam optimizer. We tune the learning rate (considering $10^{-2}$, $10^{-3}$ and $10^{-4}$ as possible values) and the mini-batch size (considering $8$, $16$ and $32$ as possible values) using grid-search.


\section{Post-retrieval Predictors}
\label{sec:Post}
All post-retrieval predictors take into account the returned list of results. Since we use the precision@$100$ to determine the ground-truth effectiveness of queries, we set $k=100$ and use the top $k$ retrieved results for the post-retrieval predictors presented below.

\subsection{Baselines}

\vspace{0.1cm}
\noindent
{\bf Score variance.}
We use the score variance introduced by Cummins et al.~\cite{Cummins-SIGIR-2011} as our first baseline post-retrieval predictor. The score variance behaves as an estimator for the influence of characteristics that are not related to the query. Because of its simplicity, this predictor can be applied to images without any adaptation. Formally, the estimated query performance is computed as follows:
\begin{equation}
s(q)=var\left(\delta\left(v_q,v_i^{(q,R)}\right)\right).
\end{equation}


\vspace{0.1cm}
\noindent
{\bf Correlation-based CNN.}
Sun et al.~\cite{Sun-TIP-2018} proposed to train a CNN on similarity or correlation matrices. A ranked list of images is turned into a matrix by computing the pairwise similarities between all pairs of returned images. The authors found that the best approach to determine the pairwise similarity between two images is to pass them through another CNN and compute the similarity between the resulting embedding vectors. To obtain these embedding vectors, we employ the embedding model $f$ which comes with the retrieval system. In our case, the model $f$ is a ResNet-101 \cite{He-CVPR-2016}. 

The CNN that learns to predict query performance based on similarity matrices is formed of three convolutional-pooling blocks, followed by two dense layers. We use the exact same configuration for the CNN as Sun et al.~\cite{Sun-TIP-2018}. The model is optimized to minimize the mean squared error between the ground-truth and the predicted query performance. Following Sun et al.~\cite{Sun-TIP-2018}, we train the CNN for $100$ epochs with the Adam optimizer. We tune the learning rate (considering $10^{-2}$, $10^{-3}$ and $10^{-4}$ as possible values) and the mini-batch size (considering $8$, $16$ and $32$ as possible values) using grid-search.

\subsection{Proposed}

\vspace{0.1cm}
\noindent
{\bf Adapted query feedback.}
Our first post-retrieval predictor is a redesigned version of the query feedback proposed by Zhou et al.~\cite{Zhou-SIGIR-2007}. In text retrieval, the query feedback is computed as the overlap between the ranked lists retrieved by a system for the original query and the expanded query, respectively. The query expansion is performed considering the list of documents retrieved for the original query. In the image domain, we perform query expansion by finding the median image in the list $\rho_{q,R}$ returned by the system $R$ for the query $q$. We define the median image as the image $q' \in \rho_{q,R}$ having the closest embedding to the average embedding of the returned list:
\begin{equation}
q'=\argmin_{x_i^{(q,R)} \in \rho_{q,R}} \delta\left(f\left(x_i^{(q,R)}\right), \frac{1}{k}\sum_{j=1}^k f\left(x_j^{(q,R)}\right)\right),    
\end{equation}
where $f$ is the embedding model.
Finally, our adapted query feedback is given by the IoU ratio between $\rho_{q,R}$ and $\rho_{q',R}$.

\vspace{0.1cm}
\noindent
{\bf Iterative removal of discriminative features.}
We propose an approach for QPP inspired by the unmasking technique of Koppel et al.~\cite{Koppel-JMLR-2007}. The technique was initially proposed for the task of text author identification, serving as an estimator for how distinguishable texts are from each other.
The method relies on gradually removing distinguishable features learned by a linear classifier,
leveraging the idea that if two texts are written by the same author, then they should differ by a relatively small amount of features.

We adapt the aforementioned principle with the purpose of identifying the level of similarity between the query and the retrieved images. 
We take the top $k$ retrieved images and compute the Hadamard product between their embeddings and the query embedding $v_q$ to identify the features with higher correlation. We sort the features in descending order of the correlation and remove the top $m$ features from the embeddings of the query and the database images. This process is repeated $l$ times. To measure query performance, we employ the IoU score computed over the sets of images retrieved at all iterations. Intuitively, if a query is easy to handle, systematic removals of features should not strongly deter the original answer, as the query exhibits a larger set of highly correlated features. We employ grid-search to perform hyperparameter tuning, obtaining optimal results with $m =50$ and $l=15$.

\vspace{0.1cm}
\noindent
{\bf Embedding variance.} 
Inspired by the intuition behind the predictor based on score variance \cite{Shtok-TIS-2012}, we propose a predictor based on estimating the variance of the embeddings of the retrieved images. Formally, the effectiveness of query $q$ is estimated by:
\begin{equation}\label{eq_embedding_var}
 s(q) = var(f(\rho_{q,R})),   
\end{equation}
where $f$ is the embedding model and $\rho_{q,R}$ is the list of $k$ images returned by the system $R$, as defined in Section~\ref{sec_formulation}. From another perspective, we can regard the predictor defined in Eq.~\eqref{eq_embedding_var} as a post-retrieval version of the pre-retrieval predictor based on k-means cluster density defined in Eq.~\eqref{eq_kmeans}. Indeed, the returned images $\rho_{q,R}$ are likely to have a high overlap with the images in cluster $C_j$. The overlap is higher, as the query image is closer to the cluster centroid. 

\vspace{0.1cm}
\noindent
{\bf Meta-regressor.}
We propose a meta-regression model to leverage the results of all of the previously described methods. We first normalize the values of all the other predictors between $0$ and $1$. Next, we train a Support Vector Regression (SVR) model and employ grid-search to identify the optimal values of the penalty term $C\in \{0.1, 1,10,100\}$, the fraction of support vectors $\nu \in \{0.1,0.25,0.5,0.75\}$, and the kernel type (linear or RBF). We identify $C=100$, $\nu = 0.25$ and the Radial Basis Function (RBF) kernel as the optimal choices. 

\section{Benchmark Resources} 

\begin{table}[t]
    \caption{Important statistics about the data sets included in the iQPP benchmark.\vspace{-0.3cm}}
    \label{tab_data}
  \begin{center}
  \begin{tabular}{lccc}
    \toprule
    Data set & \#images     & \#train queries          &  \#test queries \\
    \midrule
    PASCAL VOC 2012 & 17,125 & 700 & 700\\
    Caltech-101 & 9,146 & 700 & 700\\
    ROxford5k & 5,063 & - & 70\\
    RParis6k & 6,392 & - & 70\\
    \bottomrule
  \end{tabular}
  \end{center}
\end{table}

\begin{figure}[t]
  \centering
  \includegraphics[width=\linewidth]{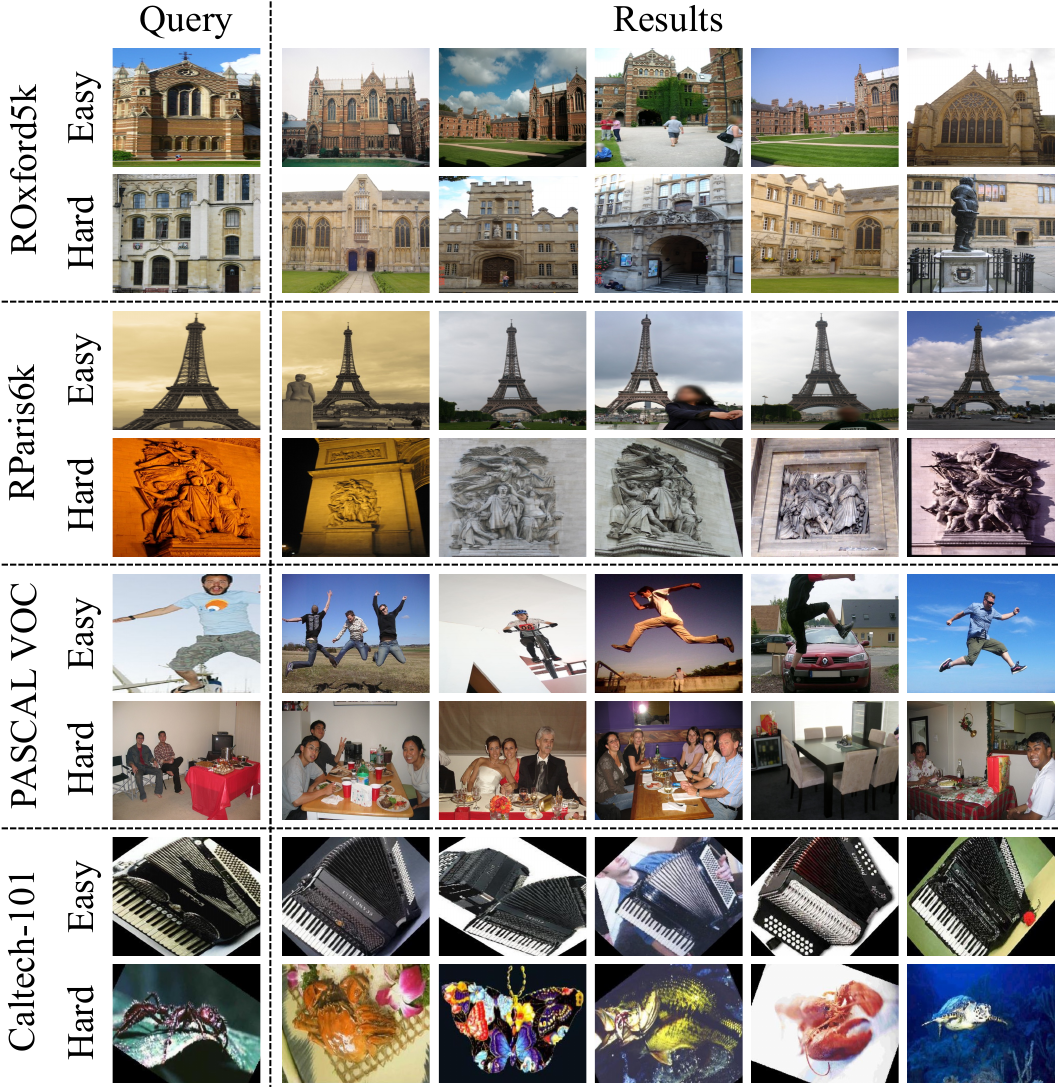}
  \vspace{-0.6cm}
  \caption{Examples of easy (high performance) and hard (low performance) queries from the four data sets included in the iQPP benchmark. For each query, we show the top five results returned by the system of Radenovi\'{c} et al.~\cite{Radenovic-TPAMI-2019} to better illustrate the performance levels of the chosen queries. Best viewed in color.}
  \Description{Examples of easy (high performance) and hard (low performance) queries from the four data sets included in the iQPP benchmark. For each query, we show the top five results returned by the system of Radenovi\'{c} et al.~\cite{Radenovic-TPAMI-2019} to better illustrate the performance levels of the chosen queries.}
  \label{fig_examples}
\end{figure}
 
\subsection{Data sets}

To evaluate the performance of the predictors, we assemble a lineup of four image data sets to form our novel benchmark, namely PASCAL VOC 2012 \cite{Everingham-IJCV-2015}, Caltech-101 \cite{Li-CT-2022}, ROxford5k \cite{Radenovic-CVPR-2018} and RParis6k \cite{Radenovic-CVPR-2018}. While ROxford5k and RParis6k are well established CBIR data sets, we repurpose PASCAL VOC 2012 and Caltech-101 to accommodate the QPP task. The well-established ROxford5k and RParis6k data sets consist of 70 queries each. To achieve a more robust evaluation on PASCAL VOC 2012 and Caltech-101, we establish a set of 700 test queries for these two data sets, increasing the number of queries by an order of magnitude compared to the well-established ROxford5k and RParis6k, as shown in Table \ref{tab_data}.

In Figure \ref{fig_examples}, we illustrate two queries (one easy and one hard) per data set, and the top five results for each query. Note that ROxford5k and RParis6k contain landmark images, while PASCAL VOC 2012 and Caltech-101 contain images of various objects.

\vspace{0.1cm}
\noindent
{\bf PASCAL VOC 2012.}
PASCAL VOC 2012 \cite{Everingham-IJCV-2015} is a data set of 17,125 images covering a wide range of computer vision tasks such as image segmentation, object detection and object recognition. It contains a broad array of real-life scenes, depicting more than 20 object categories. The data set provides annotations for bounding boxes, object classes and contours. 

Our work increases the value of the data set by supplementing the available tasks with an additional entry, that of content-based image retrieval. We generate a selection of query images which are held out from the image list. We consider two variations of the retrieval process: easy and hard. For the easy track, we crop the bounding boxes of random objects and use the cropped objects as query images, marking as positive any image that contains the searched object class. The difficult track focuses on multi-object queries (illustrating certain activities,  e.g.~person riding a bike, cats playing), marking as positive any image that contains all of the object classes seen in the query image. We generate 350 single-object queries and 350 multi-object queries for evaluation purposes. For supervised predictors, we generate an equally large and identically balanced set of queries (700) for training.


\vspace{0.1cm}
\noindent
{\bf Caltech-101.}
Caltech-101 \cite{Li-CT-2022} is an object recognition data set, which we repurpose to address the CBIR task. The data set consists of images depicting objects from one of 100 object classes. Additional images are provided to represent a visual clutter (background) class.
The data set contains 9,146 images. We keep 700 images as training queries and another 700 images as test queries.

\vspace{0.1cm}
\noindent
{\bf ROxford5k.}
ROxford5k \cite{Radenovic-CVPR-2018} is a version of the Oxford5k \cite{Philbin-CVPR-2007} data set curated by Radenovi\'{c} et al.~\cite{Radenovic-CVPR-2018}. Oxford5k is a popular object retrieval data set that contains images depicting landmarks from the city of Oxford. The curation consists of annotation corrections, new queries and multiple difficulty tracks. ROxford5k is composed of 5,063 images, out of which 70 are held out as queries. The data samples are split into four different categories based on how clearly they depict the subject: easy, hard, negative and unclear. Images labeled as easy have minor viewpoint changes and illumination conditions similar to the query, while images labeled as hard have more difficult viewing conditions. Unclear images depict landmarks that cannot be accurately identified without contextual information. Since ROxford5k \cite{Radenovic-CVPR-2018} comes with only 70 queries, we propose to evaluate supervised predictors using a 5-fold cross-validation procedure. We publicly release the folds with the benchmark.


\vspace{0.1cm}
\noindent
{\bf RParis6k.}
RParis6k \cite{Radenovic-CVPR-2018} is an enhancement of Paris6k \cite{Philbin-CVPR-2008} consisting of 6,392
images and 70 queries. The data set follows an identical structure to ROxford5k, since it is also curated by Radenovi\'{c} et al.~\cite{Radenovic-CVPR-2018}. RParis6k contains various landmarks from the city of Paris from multiple viewing points and illumination conditions. To evaluate supervised query performance predictors, we employ 5-fold cross-validation. As for ROxford5k, we make the folds public to facilitate future comparisons.


\subsection{Evaluation protocol}

To assess the difficulty level of a query, we consider two alternative measures of retrieval effectiveness, namely the average precision (AP) and the precision for the top $k$ retrieved results (P@$k$). The precision@$k$ is given by the number of true positive images divided by $k$. The recall is given by the ratio between the number of true positive images and the number of images labeled as positive for the query. The AP is given by the area under the precision-recall curve, which takes into account all possible thresholds $k$. Although P@$10$ is sometimes used in text QPP \cite{YomTov-SIGIR-2005}, we found that a high percentage of the test queries (between $29\%$ and $82\%$, depending on the data set) have a P@$10$ score of $1$. For a better estimation of query difficulty, we decided to use P@$100$.

To estimate the performance level of a predictor, we employ the Pearson and Kendall $\tau$ correlation coefficients between the predicted and the actual effectiveness levels of all test queries, following the conventional evaluation procedure in text QPP~\cite{YomTov-SIGIR-2005,Zhao-ECIR-2008,Chifu-SIGIR-2018,Faggioli-IRJ-2022}. Moreover, we apply a Student's t-test at a confidence score of $0.01$ to test significance~\cite{Roitman-SIGIR-2018}.
Although some data sets separate queries into different difficulty tracks, we aim to evaluate the capacity of predictors to estimate the actual AP or P@100 scores rather than classifying the queries as easy or hard, since we consider that the regression task is more suitable for revealing the true abilities of the predictors. 

\begin{table*}[t]
\caption{Pearson and Kendall $\mathbf{\tau}$ correlations of the query performance predictors on PASCAL VOC 2012 and Caltech-101, which contain images of various object classes. Underlined results are significantly better than the random chance baseline, according to a Student's t-test with a p-value lower than $\mathbf{0.01}$. The top pre-retrieval and post-retrieval scores are highlighted in bold.\vspace{-0.3cm}}
    \label{tab_results_pascaltech}
\centering
\setlength{\tabcolsep}{2.0pt}
\begin{tabular}{lllrrrrrrrrrrrrrrrr}
\toprule
\multirow{6}{*}{\rotatebox{90}{Type}} & \multirow{6}{*}{\rotatebox{90}{Supervised}} & \multirow{6}{*}{Method} & \multicolumn{8}{c}{PASCAL VOC 2012} & \multicolumn{8}{c}{Caltech-101}\\
 & &  & \multicolumn{4}{c}{Radenovi\'{c} et al.~\cite{Radenovic-TPAMI-2019}} & \multicolumn{4}{c}{Revaud et al.~\cite{Revaud-ICCV-2019}}  & \multicolumn{4}{c}{Radenovi\'{c} et al.~\cite{Radenovic-TPAMI-2019}} & \multicolumn{4}{c}{Revaud et al.~\cite{Revaud-ICCV-2019}} \\
 & &  & \multicolumn{2}{c}{AP} & \multicolumn{2}{c}{P@100} & \multicolumn{2}{c}{AP} & \multicolumn{2}{c}{P@100} & \multicolumn{2}{c}{AP} & \multicolumn{2}{c}{P@100} & \multicolumn{2}{c}{AP} & \multicolumn{2}{c}{P@100} \\
& & & \multicolumn{1}{c}{\rotatebox{90}{Pearson\;}} & \multicolumn{1}{c}{\rotatebox{90}{Kendall}} & \multicolumn{1}{c}{\rotatebox{90}{Pearson}} & \multicolumn{1}{c}{\rotatebox{90}{Kendall}} & \multicolumn{1}{c}{\rotatebox{90}{Pearson}} & \multicolumn{1}{c}{\rotatebox{90}{Kendall}} & \multicolumn{1}{c}{\rotatebox{90}{Pearson}} & \multicolumn{1}{c}{\rotatebox{90}{Kendall}} & 
\multicolumn{1}{c}{\rotatebox{90}{Pearson}} & \multicolumn{1}{c}{\rotatebox{90}{Kendall}} &
\multicolumn{1}{c}{\rotatebox{90}{Pearson}} & \multicolumn{1}{c}{\rotatebox{90}{Kendall}} &
\multicolumn{1}{c}{\rotatebox{90}{Pearson}} & \multicolumn{1}{c}{\rotatebox{90}{Kendall}} &
\multicolumn{1}{c}{\rotatebox{90}{Pearson}} & \multicolumn{1}{c}{\rotatebox{90}{Kendall}} 
\\
\midrule
& & Random & $0.00$ & $0.00$ & $0.00$ & $0.00$ & $0.00$ & $0.00$ & $0.00$  & $0.00$ & $0.00$ & $0.00$ & $0.00$ & $0.00$ & $0.00$ & $0.00$ & $0.00$  & $0.00$  \\
\midrule
\multirow{8}{*}{\rotatebox{90}{Pre-retrieval}} 
& & \#objects / area \cite{Soviany-CVIU-2021} & $0.02$ & $\ul{\textbf{0.22}}$ & $0.03$ & $\ul{\textbf{0.25}}$ & $0.02$ & $\textbf{0.27}$ & $0.03$ & $\ul{\textbf{0.25}}$ & $0.01$ & \ul{$0.08$} & $0.01$ & \ul{$0.04$} & $0.04$ & \ul{$0.06$} & $0.03$ & $0.04$\\
& & Image difficulty \cite{Ionescu-CVPR-2016} & $\ul{\textbf{0.25}}$ & \ul{$0.19$} & $\ul{\textbf{0.33}}$ & \ul{$0.23$} & $\ul{\textbf{0.32}}$ & \ul{$0.24$} & $\ul{\textbf{0.31}}$ & \ul{$0.22$} & $-0.01$ & $-0.02$ & $-0.07$ & $-0.07$ & $0.00$ & $-0.02$ & $-0.07$ & $-0.06$ \\
& & Denoising AE & \ul{$0.15$} & \ul{$0.16$} & $0.06$ & \ul{$0.08$} & \ul{$0.11$} & \ul{$0.12$} & $0.08$ & \ul{$0.09$} & $0.03$ & $0.02$ & $0.06$ & $0.03$ & $0.12$ & $0.07$ & \ul{$0.13$} & \ul{$0.07$}\\
& & Masked AE & \ul{$0.11$} & \ul{$0.11$} & $0.01$ & $0.05$ & $0.01$ & $0.06$ & $-0.01$ & $0.03$ & $-0.04$ & $-0.04$ & $0.01$ & $0.00$ & $0.03$ & $0.02$ & $0.09$ & $0.05$\\
& & Class head kurtosis & $0.05$ & \ul{$0.08$} & $0.09$ & \ul{$0.07$} & \ul{$0.12$} & \ul{$0.09$} & \ul{$0.12$} & \ul{$0.08$} & \ul{$0.16$} & \ul{$0.17$} & \ul{$0.26$} & \ul{$0.30$} & \ul{$0.23$} & \ul{$0.17$} & \ul{$0.13$} & \ul{$0.10$} \\
& & Class head dispersion & $0.08$ & \ul{$0.09$} & \ul{$0.13$} & \ul{$0.08$} & \ul{$0.17$} & \ul{$0.11$} & \ul{$0.17$} & \ul{$0.10$} & \ul{$0.25$} & \ul{$0.20$} & \ul{$\textbf{0.48}$} & \ul{$\textbf{0.38}$} & \ul{$0.32$} & \ul{$0.23$} & \ul{$0.21$} & \ul{$0.15$} \\
& & Cluster density & \ul{$0.13$}& \ul{$0.12$} & $0.00$ & $0.01$ & $-0.02$ & $-0.04$ & $-0.01$ & $-0.01$ & \ul{$0.15$} & $0.09$ & \ul{$0.41$} & \ul{$0.24$} & $-0.13$ & \ul{$0.09$} & \ul{$-0.03$} & \ul{$-0.4$}\\
& \checkmark & Fine-tuned ViT & $0.04$ & $0.02$ & \ul{$0.20$} & \ul{$0.10$} &  \ul{$0.17$} & $0.06$ & \ul{$0.14$} & $0.05$ & $\ul{\textbf{0.54}}$ & $\ul{\textbf{0.38}}$ & \ul{$0.27$} & \ul{$0.15$} & $\ul{\textbf{0.65}}$ & \ul{$\textbf{0.47}$} & \ul{$\textbf{0.41}$} & \ul{$\textbf{0.20}$} \\
\midrule
\multirow{6}{*}{\rotatebox{90}{Post-retrieval}} 
& & Score Variance \cite{Cummins-SIGIR-2011} & $0.02$ & $0.05$ & $-0.02$ & $0.02$ & \ul{$0.23$} & \ul{$0.19$} & \ul{$0.26$} & \ul{$0.20$} & \ul{$0.11$} & $0.01$ & \ul{$0.21$} & $0.01$ & \ul{$0.51$} & \ul{$0.51$} & \ul{$0.30$} & \ul{$0.39$}\\
& \checkmark & Correlation CNN \cite{Sun-TIP-2018} & \ul{$0.27$} & $0.07$  & \ul{$0.32$} & \ul{$0.16$} & \ul{$0.32$} & \ul{$0.15$} &\ul{$0.26$} & \ul{$0.11$} & \ul{$\textbf{0.83}$} & \ul{$\textbf{0.65}$}  & \ul{$\textbf{0.76}$} & \ul{$\textbf{0.51}$} & \ul{$\textbf{0.78}$} & \ul{$\textbf{0.60}$} & \ul{$\textbf{0.71}$} & \ul{$\textbf{0.50}$}\\
& & Adapted query feedback & \ul{$0.23$} & \ul{$0.16$} & \ul{$0.37$} & \ul{$0.21$} & \ul{$0.41$} & \ul{$0.26$} & \ul{$0.41$} & \ul{$0.24$} & \ul{$0.60$}& \ul{$0.43$} & \ul{$0.60$} & \ul{$0.46$} & \ul{$0.56$} & \ul{$0.40$} & \ul{$0.60$} & \ul{$0.44$}\\
& & Iterative removal & \ul{$0.16$} & \ul{$0.13$} & \ul{$0.35$} & \ul{$0.20$} & \ul{$0.41$} & \ul{$0.26$} & \ul{$0.40$} & \ul{$0.23$} & 
\ul{$0.57$} & \ul{$0.41$} & \ul{$0.57$} & \ul{$0.42$} & \ul{$0.31$} & \ul{$0.20$} & \ul{$0.40$} & \ul{$0.23$} \\
& & Embedding Variance & \ul{$0.29$} & \ul{$0.20$} & \ul{$0.33$} & \ul{$0.21$} & \ul{$0.43$} & \ul{$0.22$} & \ul{$0.37$} & \ul{$0.20$} & \ul{$0.28$} & \ul{$0.20$} & \ul{$0.49$} & \ul{$0.28$} & \ul{$0.26$} & \ul{$0.18$} & \ul{$0.49$} & \ul{$0.26$}\\
& \checkmark & Meta-regressor & \ul{$\textbf{0.36}$} & \ul{$\textbf{0.28}$} & \ul{$\textbf{0.45}$} & \ul{$\textbf{0.29}$} & \ul{$\textbf{0.51}$} & \ul{$\textbf{0.34}$} & \ul{$\textbf{0.48}$} & \ul{$\textbf{0.30}$} & \ul{$0.71$} & \ul{$0.53$} & \ul{$0.72$} & \ul{$\textbf{0.51}$} & \ul{$0.76$} & \ul{$0.57$} & \ul{$0.70$} & \ul{$0.49$}\\
\bottomrule
\end{tabular}
\end{table*}

\subsection{Image retrieval models}

The first image retrieval system used in our benchmark was proposed by Radenovi\'{c} et al.~\cite{Radenovic-TPAMI-2019}\footnote{\url{https://github.com/filipradenovic/cnnimageretrieval-pytorch}}. The model is based on fine-tuning a convolutional neural network on a large set of annotation-free images. The authors leverage the use of geometry and camera positioning of 3D models returned by a structure-from-motion framework to guide the selection of matching and non-matching image pairs, eliminating the need for manually annotated data. The proposed architecture employs a novel pooling layer with trainable parameters that induce a particular case of the generalized mean (GeM).

The second image retrieval system from iQPP was introduced by Revaud et al.~\cite{Revaud-ICCV-2019}\footnote{\url{https://github.com/naver/deep-image-retrieval}}. The authors apply a histogram binning approximation to make the AP differentiable, enabling its use as a loss function for training deep networks. The system uses a ResNet-101 \cite{He-CVPR-2016} backbone pre-trained on ImageNet \cite{Russakovsky-IJCV-2015}. Following Radenovi\'{c} et al.~\cite{Radenovic-TPAMI-2019}, a GeM pooling layer is integrated into the architecture.

\begin{table*}[t]
\caption{Pearson and Kendall $\mathbf{\tau}$ correlations of the query performance predictors on ROxford5k and RParis6k, which contain landmark images. Underlined results are significantly better than the random chance baseline, according to a Student's t-test with a p-value lower than $\mathbf{0.01}$. The top pre-retrieval and post-retrieval scores are highlighted in bold.\vspace{-0.3cm}}
    \label{tab_results_oxford_paris}
\centering
\setlength{\tabcolsep}{1.9pt}
\begin{tabular}{lllrrrrrrrrrrrrrrrr}
\toprule
\multirow{6}{*}{\rotatebox{90}{Type}} & \multirow{6}{*}{\rotatebox{90}{Supervised}} & \multirow{6}{*}{Method} & \multicolumn{8}{c}{ROxford5k} & \multicolumn{8}{c}{RParis6k}\\
&  &  & \multicolumn{4}{c}{Radenovi\'{c} et al.~\cite{Radenovic-TPAMI-2019}} & \multicolumn{4}{c}{Revaud et al.~\cite{Revaud-ICCV-2019}}  & \multicolumn{4}{c}{Radenovi\'{c} et al.~\cite{Radenovic-TPAMI-2019}} & \multicolumn{4}{c}{Revaud et al.~\cite{Revaud-ICCV-2019}} \\
&  &  & \multicolumn{2}{c}{AP} & \multicolumn{2}{c}{P@100} & \multicolumn{2}{c}{AP} & \multicolumn{2}{c}{P@100} & \multicolumn{2}{c}{AP} & \multicolumn{2}{c}{P@100} & \multicolumn{2}{c}{AP} & \multicolumn{2}{c}{P@100} \\
& & & \multicolumn{1}{c}{\rotatebox{90}{Pearson\;}} & \multicolumn{1}{c}{\rotatebox{90}{Kendall}} & \multicolumn{1}{c}{\rotatebox{90}{Pearson}} & \multicolumn{1}{c}{\rotatebox{90}{Kendall}} & \multicolumn{1}{c}{\rotatebox{90}{Pearson}} & \multicolumn{1}{c}{\rotatebox{90}{Kendall}} & \multicolumn{1}{c}{\rotatebox{90}{Pearson}} & \multicolumn{1}{c}{\rotatebox{90}{Kendall}} & 
\multicolumn{1}{c}{\rotatebox{90}{Pearson}} & \multicolumn{1}{c}{\rotatebox{90}{Kendall}} &
\multicolumn{1}{c}{\rotatebox{90}{Pearson}} & \multicolumn{1}{c}{\rotatebox{90}{Kendall}} &
\multicolumn{1}{c}{\rotatebox{90}{Pearson}} & \multicolumn{1}{c}{\rotatebox{90}{Kendall}} &
\multicolumn{1}{c}{\rotatebox{90}{Pearson}} & \multicolumn{1}{c}{\rotatebox{90}{Kendall}} 
\\
\midrule
& & Random & $0.00$ & $0.00$ & $0.01$ & $0.00$ & $0.00$ & $0.00$ & $0.02$  & $0.01$ & $0.00$ & $0.00$ & $0.01$ & $0.00$ & $0.01$ & $0.00$ & $0.02$  & $0.01$ \\
\midrule
\multirow{8}{*}{\rotatebox{90}{Pre-retrieval}} 
& & \#objects / area \cite{Soviany-CVIU-2021} & $0.17$ & $0.05$ & $0.14$ & $0.03$ & $0.16$ & $0.02$ & $-0.05$ & $0.04$ & $0.01$ & $-0.01$ & $-0.17$ & $-0.16$ & $0.00$ & $-0.07$ & $-0.12$ & $-0.19$ \\
& & Image difficulty \cite{Ionescu-CVPR-2016} & $-0.04$ & $-0.04$ & $-0.18$ & $-0.11$ & $-0.11$ & $-0.10$ & $0.12$ & $0.09$  & $-0.06$ & $-0.03$ & $-0.04$ & $0.05$ & $-0.19$ & $-0.06$ & $-0.09$ & $-0.01$\\
& & Denoising AE & $-0.07$ & $-0.02$ & $-0.04$ & $0.06$ & $-0.07$ & $-0.03$ & $0.06$ & $0.09$  & $-0.18$ & $-0.20$ & \ul{$0.31$} & $0.19$ & $0.02$ & $0.00$ & \ul{$\textbf{0.45}$} & \ul{$0.28$}\\
& & Masked AE & $0.10$ & $0.07$ & $0.22$ & $0.20$ & $-0.09$ & $0.07$ & $0.11$ & $0.11$ & $-0.21$ & $-0.20$ & $0.18$ & $0.14$ & $0.10$ & $0.02$ & \ul{$0.40$} & \ul{$0.26$} \\
& & Class head kurtosis & $0.27$ & $0.20$ & $0.20$ & $0.18$ & $0.28$ & $0.18$ & $0.00$ & $0.04$ & $0.19$ & $0.18$ & $-0.01$ & $-0.09$ & $0.18$ & $0.00$ & $0.04$ & $-0.07$ \\
& & Class head dispersion & \ul{$0.34$} & \ul{$0.24$} & $0.27$ & \ul{$\textbf{0.23}$} & \ul{$0.35$} & \ul{$0.21$} & $0.12$ & $0.10$ & \ul{$\textbf{0.34}$} & \ul{$\textbf{0.24}$} & $0.27$ & \ul{$\textbf{0.23}$} & \ul{$\textbf{0.35}$} & \ul{$0.21$} & $0.12$ & $0.10$ \\
& & Cluster density & $0.22$ & $0.02$ & \ul{$0.38$} & $0.14$ & $0.01$ & $-0.05$ & $0.23$ & $0.01$ & $0.32$ & \ul{$0.22$}& \ul{$\textbf{0.36}$} & \ul{$0.21$} & $0.20$ & $0.17$ & $0.23$ & $0.17$ \\
& \checkmark & Fine-tuned ViT  &  \ul{$\textbf{0.41}$} & \ul{$\textbf{0.31}$}  & \ul{$\textbf{0.40}$} & \ul{$0.21$} & \ul{$\textbf{0.43}$} & \ul{$\textbf{0.28}$} & \ul{$\textbf{0.36}$} & \ul{$\textbf{0.22}$} & $0.18$ & \ul{$0.21$} & 
$0.23$ & $0.17$ & $0.25$ & \ul{$\textbf{0.24}$} & \ul{$0.38$} & \ul{$\textbf{0.38}$} \\
\midrule
\multirow{6}{*}{\rotatebox{90}{Post-retrieval}} 
& & Score variance \cite{Cummins-SIGIR-2011} & $0.01$ & $-0.02$ & $0.21$ & \ul{$0.22$} & $  0.27$ & \ul{$0.28$} & \ul{$0.41$} & \ul{$0.35$} & $0.41$ & $0.17$ & \ul{$0.45$} & \ul{$0.35$} & $-0.02$ & $0.02$ & \ul{$-0.38$} & $-0.12$ \\
& \checkmark & Correlation CNN \cite{Sun-TIP-2018} & \ul{$0.43$} & \ul{$0.30$} & \ul{$\textbf{0.69}$} & \ul{$0.49$} & \ul{$\textbf{0.60}$} & \ul{$0.44$} & \ul{$\textbf{0.90}$} & \ul{$0.62$} & \ul{$0.65$} & \ul{$0.33$} & \ul{$\textbf{0.77}$} & \ul{$0.55$} & \ul{$\textbf{0.56}$} & \ul{$\textbf{0.46}$} & \ul{$\textbf{0.66}$} & \ul{$0.37$}\\
& & Adapted query feedback & $0.22$ & $0.11$ & \ul{$0.36$} & \ul{$0.24$} & \ul{$0.42$} & \ul{$0.29$} & \ul{$0.77$} & \ul{$0.52$} & \ul{$0.33$} & $0.13$ & \ul{$0.56$} & \ul{$0.32$} & \ul{$0.32$} & $0.16$ & \ul{$0.32$} & \ul{$0.26$} \\
& & Iterative removal & $0.28$ & \ul{$0.22$} & \ul{$0.31$} & \ul{$0.33$} & \ul{$0.36$} & \ul{$0.23$} & \ul{$0.75$} & \ul{$0.53$} & $0.30$ & $0.17$ & \ul{$0.40$} & $0.19$ & \ul{$0.35$} & $0.14$ & \ul{$0.50$} & \ul{$0.32$}\\
& & Embedding variance & \ul{$0.30$} & $0.15$ & \ul{$0.54$} & \ul{$0.32$} & \ul{$0.34$} & \ul{$0.23$} & 
\ul{$0.84$} & \ul{$0.57$} & \ul{$0.51$} & \ul{$0.27$} &  \ul{$0.69$} & \ul{$0.45$} &  \ul{$0.36$} & \ul{$0.18$} & \ul{$0.47$} & \ul{$0.40$} \\
& \checkmark & Meta-regressor & \ul{$\textbf{0.49}$} & \ul{$\textbf{0.37}$} & \ul{$0.62$} & \ul{$\textbf{0.51}$} & \ul{$0.58$} & \ul{$\textbf{0.45}$} & \ul{$0.88$} & \ul{$\textbf{0.65}$} & \ul{$\textbf{0.69}$} & \ul{$\textbf{0.51}$} & \ul{$0.76$} & \ul{$\textbf{0.60}$} & \ul{$0.37$} & \ul{$0.37$} & \ul{$0.65$} &\ul{$\textbf{0.56}$} \\
\bottomrule
\end{tabular}
\end{table*}

\section{Benchmark Results}

We group our experiments based on the type of images from the chosen data sets. Hence, in Section~\ref{sec_results_objects}, we present the results on data sets composed of images of various natural or man-made objects, namely PASCAL VOC 2012 and Caltech-101. In Section~\ref{sec_results_landmark}, we discuss the results on ROxford5k and RParis6k, as both data sets contain landmark images. Finally, we make a few observations about the overall results in Section~\ref{sec_results_overall}.

\subsection{Results on PASCAL VOC and Caltech-101}
\label{sec_results_objects} 

We report the results on PASCAL VOC 2012 and Caltech-101 in Table \ref{tab_results_pascaltech}. We discuss the reported results below, making several interesting observations. 

\vspace{0.1cm}
\noindent
{\bf Results of pre-retrieval predictors.}
On PASCAL VOC, the best pre-retrieval predictors are the baselines based on image difficulty or the number of objects divided by their area (see the first two rows after the random baseline in Table~\ref{tab_results_pascaltech}). These pre-retrieval predictors mainly rely on the presence of multiple objects from various object categories inside the query image, e.g.~people, cars or dogs, being suitable for PASCAL VOC queries. However, the two baseline predictors exhibit poor performance on Caltech-101 images, which typically contain only one object per image. Interestingly, on PASCAL VOC, image difficulty gives a better Pearson correlation, while the number of objects divided by their area gives a better Kendall $\tau$. This observation points towards the importance of using multiple correlation measures to better assess predictors' behavior.

On Caltech-101, the best pre-retrieval predictor for the AP measure is the fine-tuned ViT. However, when we consider P@$100$ as the ground-truth query performance, the fine-tuned ViT is surpassed by the classification head dispersion in one scenario. Some predictors seem to be better suited for certain effectiveness measures. To find robust predictors across effectiveness measures, it is thus important to include more than a single measure to estimate ground-truth query performance.
Finally, we observe that the two data sets have distinct top scoring pre-retrieval predictors, demonstrating that it is not sufficient to use a single data set to find generic predictors.

\begin{figure}[t]
  \centering
  \includegraphics[width=\linewidth]{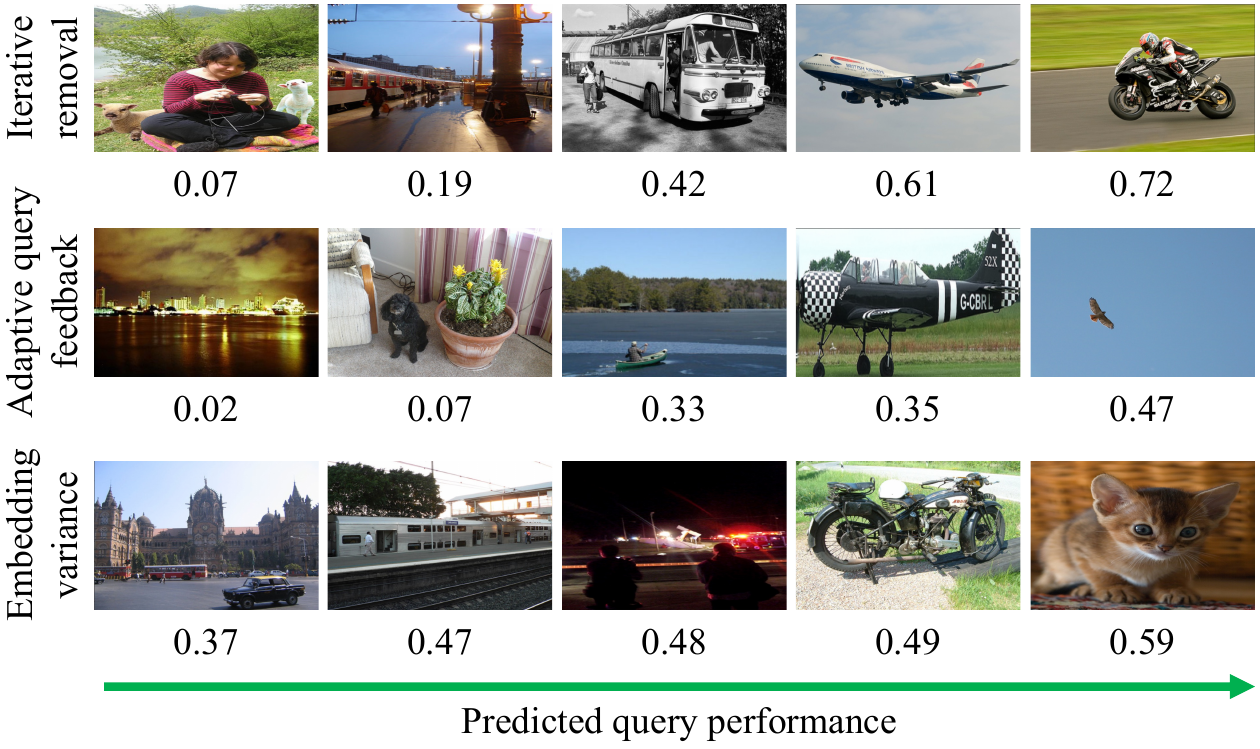}
  \vspace{-0.6cm}
  \caption{Image queries from PASCAL VOC 2012 displayed in increasing order of predicted performance, from left to right. Examples are illustrated for three of the best unsupervised predictors: iterative feature removal (top row), adaptive query feedback (middle row), and embedding variance (bottom row). Best viewed in color.}
  \Description{Image queries from PASCAL VOC 2012 displayed in increasing order of predicted performance, from left to right. Examples are illustrated for three of the best unsupervised predictors: iterative feature removal (top row), adaptive query feedback (middle row), and embedding variance (bottom row).}
  \label{fig_predicted_qp}
  \vspace{-0.3cm}
\end{figure}

\vspace{0.1cm}
\noindent
{\bf Results of post-retrieval predictors.}
The baseline predictor based on score variance seems to be a suitable estimator for the system of Revaud et al.~\cite{Revaud-ICCV-2019}. However, the predictions given by score variance for the system of Radenovi\'{c} et al.~\cite{Radenovic-TPAMI-2019} are close to random chance. These results demonstrate that predictors can depend on the reference system. By including multiple retrieval systems in our benchmark, we are able to identify predictors that are inconsistent across different retrieval models.

On PASCAL VOC 2012, our largest data set, the meta-regressor outperforms all competing predictors, leveraging the use of information from the other predictors to surpass them. However, this does not happen on Caltech-101, where the best predictor is the correlation-based CNN. Regardless of the data set, it is clear that supervised post-retrieval predictors are generally better, surpassing the unsupervised post-retrieval predictors. Another expected outcome is that post-retrieval predictors obtain superior results compared with pre-retrieval predictors.

The proposed unsupervised post-retrieval predictors (adapted query feedback, iterative feature removal and embedding variance) reach reasonably good correlation levels, always surpassing the random predictor baseline by statistically significant margins. To further analyze the behavior of these predictors, we illustrate some randomly chosen queries from PASCAL VOC 2012 in Figure \ref{fig_predicted_qp}, organizing them in increasing order of predicted performance. The figure shows that all three predictors find query images with fewer objects and plain background as more likely to exhibit high performance. In contrast, images with multiple objects, photographed in poor illumination conditions are associated with low performance levels. In summary, we find strong connections between the visual content of queries and the performance scores predicted by the unsupervised post-retrieval predictors.

\subsection{Results on ROxford5k and RParis6k}
\label{sec_results_landmark} 

We report the results on ROxford5k and RParis6k in Table \ref{tab_results_oxford_paris}. 

\vspace{0.1cm}
\noindent
{\bf Results of pre-retrieval predictors.}
Since the ROxford5k and RParis6k data sets contain landmark images, the predictors based on image difficulty and the number of objects divided by their area obtain generally poor performance. This happens because the images contain landmarks, such as the Eiffel Tower, rather than objects, such as dogs and horses.

The denoising and masked auto-encoders are better correlated with P@$100$ than with the AP measure, likely because P@$100$ associates higher penalties to queries with very few positive results. Since these queries likely reside in a sparse area of the data distribution (due to the low number of similar images), AE models are unable to reconstruct the corresponding queries very well. This observation shows the importance of using more than one ground-truth query performance metric to find predictors that are robust across multiple target performance measures, supporting our decision to use both the AP and P@$100$ measures for our benchmark. 

Among the pre-retrieval predictors, the fine-tuned ViT obtains the best results in most cases. This is rather unsurprising, since ViT is a supervised predictor, while all the other pre-retrieval predictors are unsupervised. Interestingly, the ViT model is often challenged by the classification head dispersion. We thus consider the latter model as the best unsupervised pre-retrieval predictor on the ROxford5k and RParis6k data sets.

\vspace{0.1cm}
\noindent
{\bf Results of post-retrieval predictors.}
The post-retrieval predictor based on score variance obtains inconsistent results, being the weakest post-retrieval predictor. Except for the score variance, the post-retrieval predictors surpass the pre-retrieval ones in the majority of cases. Comparing the unsupervised post-retrieval predictors among each other, we observe that embedding variance provides the best scores, generally surpassing the iterative feature removal and the adapted query feedback, respectively.

The best post-retrieval predictors are the supervised ones, namely the correlation-based CNN and the meta-regressor. In 9 out 16 cases, the meta-regressor obtains the best correlations. In the other 7 cases, the correlation-based CNN outperforms all competing predictors. Interestingly, we observe that there are 5 situations where the correlation-based CNN gives a better Pearson correlation than the meta-regressor, while the meta-regressor surpasses the  correlation-based CNN in terms of Kendall $\tau$. This shows the importance of using multiple measures to evaluate QPP methods, indicating that our decision to consider both Pearson and Kendall $\tau$ for iQPP is useful in finding predictors that are consistent across QPP evaluation measures.

\begin{figure}[!t]
\centering
\subfloat[Correlations for PASCAL VOC.]
{	
    \hspace{-0.2cm}
	\includegraphics[width=0.48\linewidth]{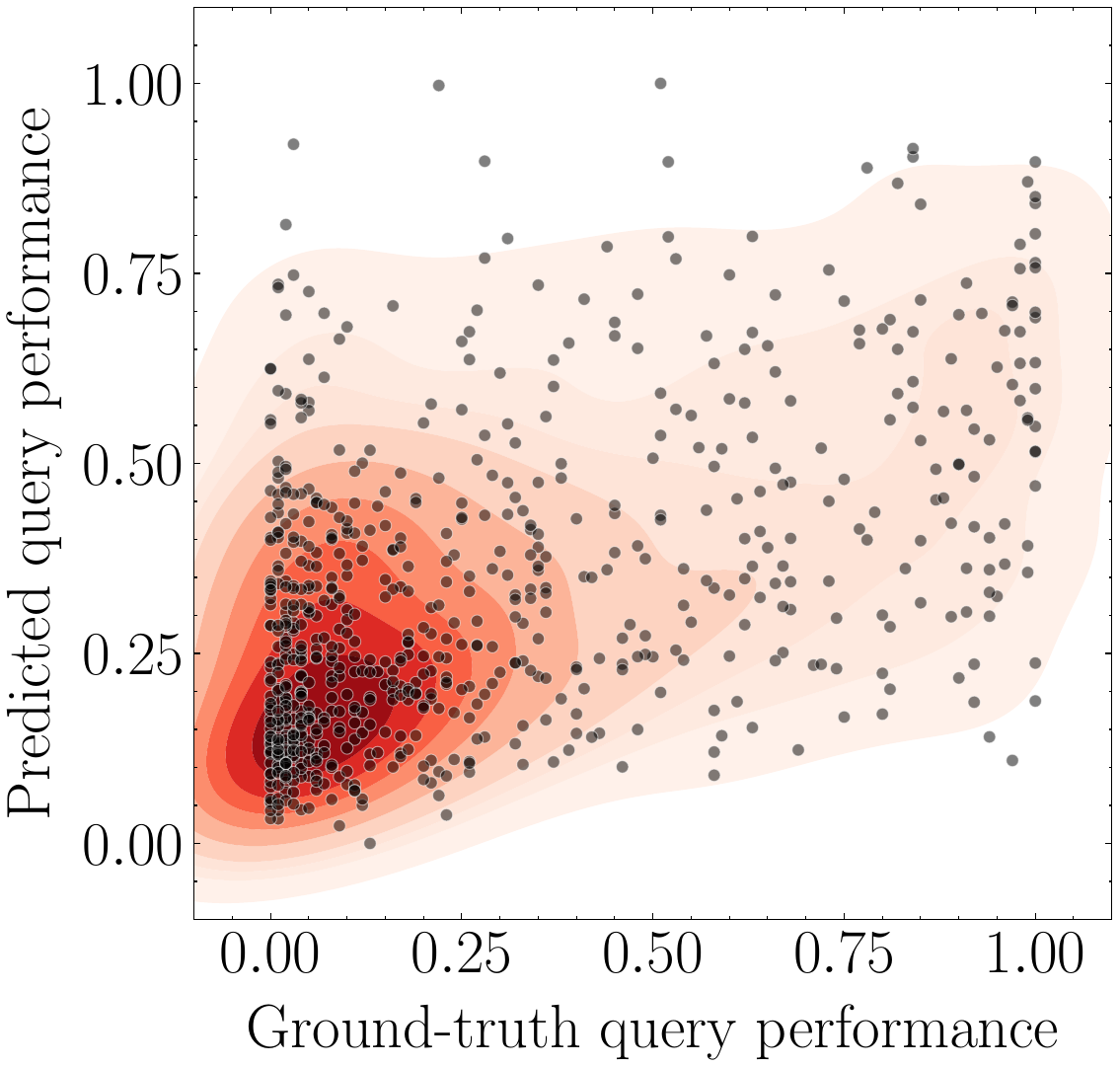}
}
\subfloat[Correlations for Caltech-101.]
{
	\includegraphics[width=0.48\linewidth]{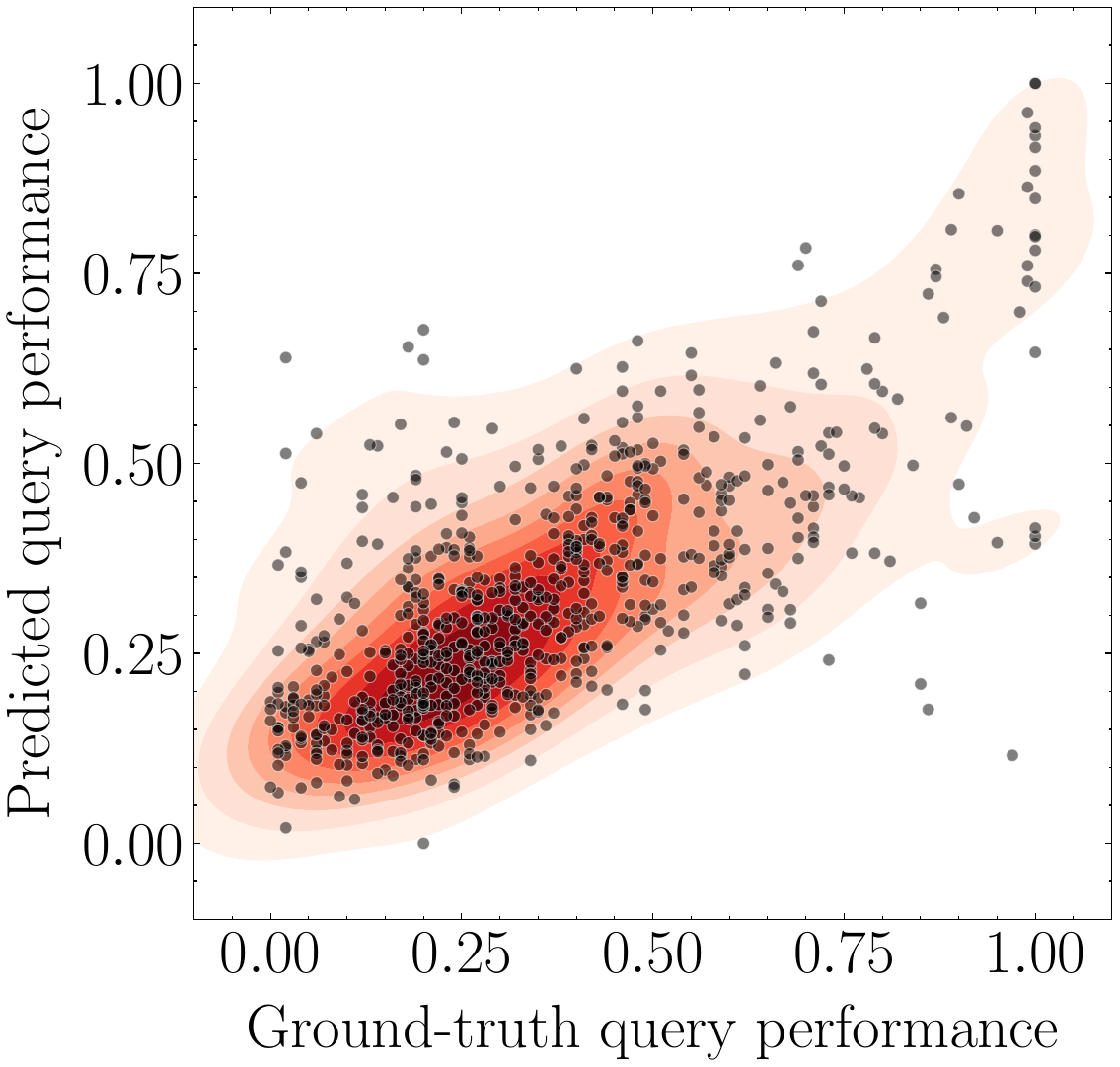}
}
\\
\vspace{-0.1cm}
\subfloat[Correlations for ROxford5k.]
{	
    \hspace{-0.2cm}
	\includegraphics[width=0.48\linewidth]{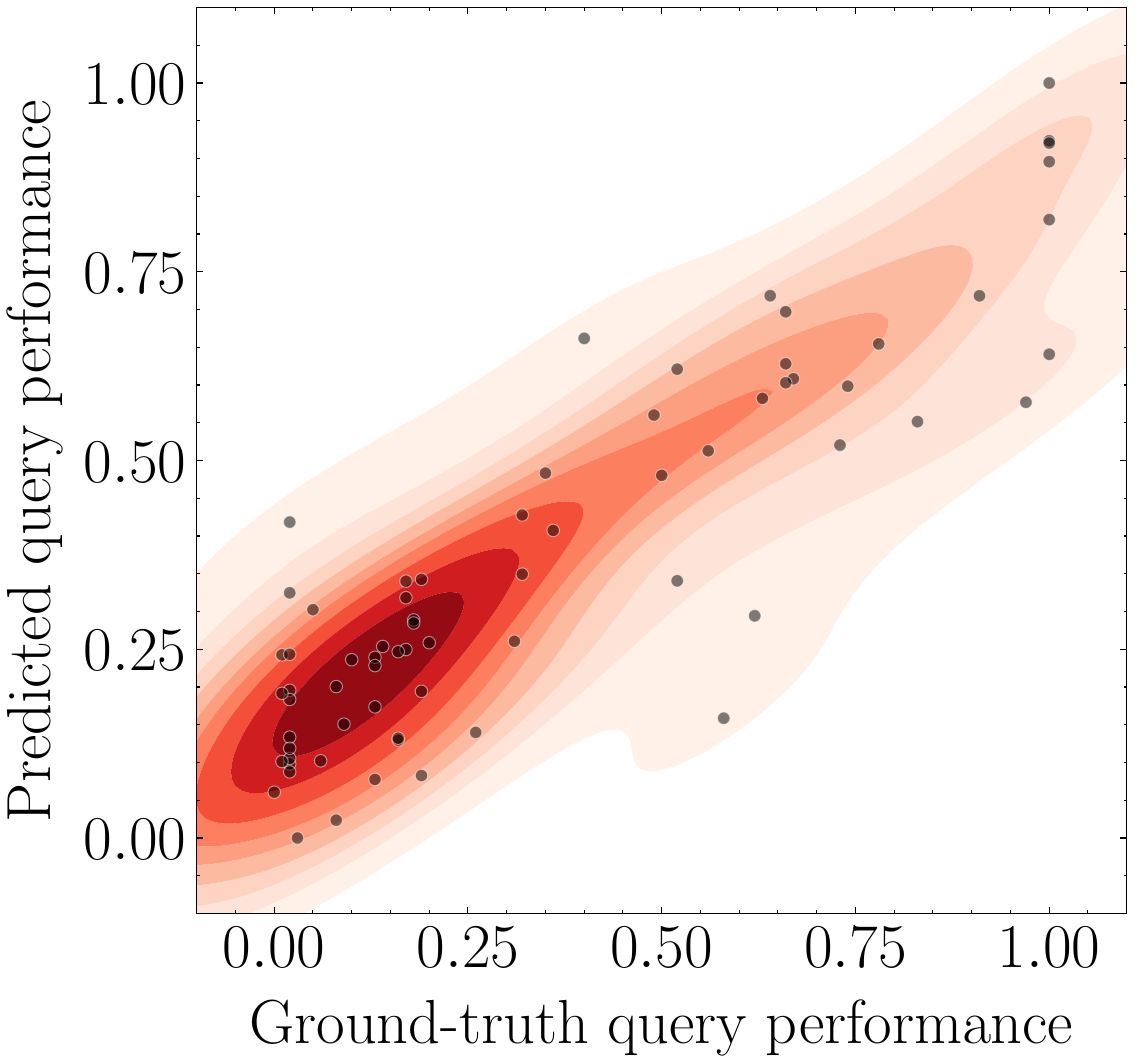}
}
\subfloat[Correlations for RParis6k.]
{
	\includegraphics[width=0.48\linewidth]{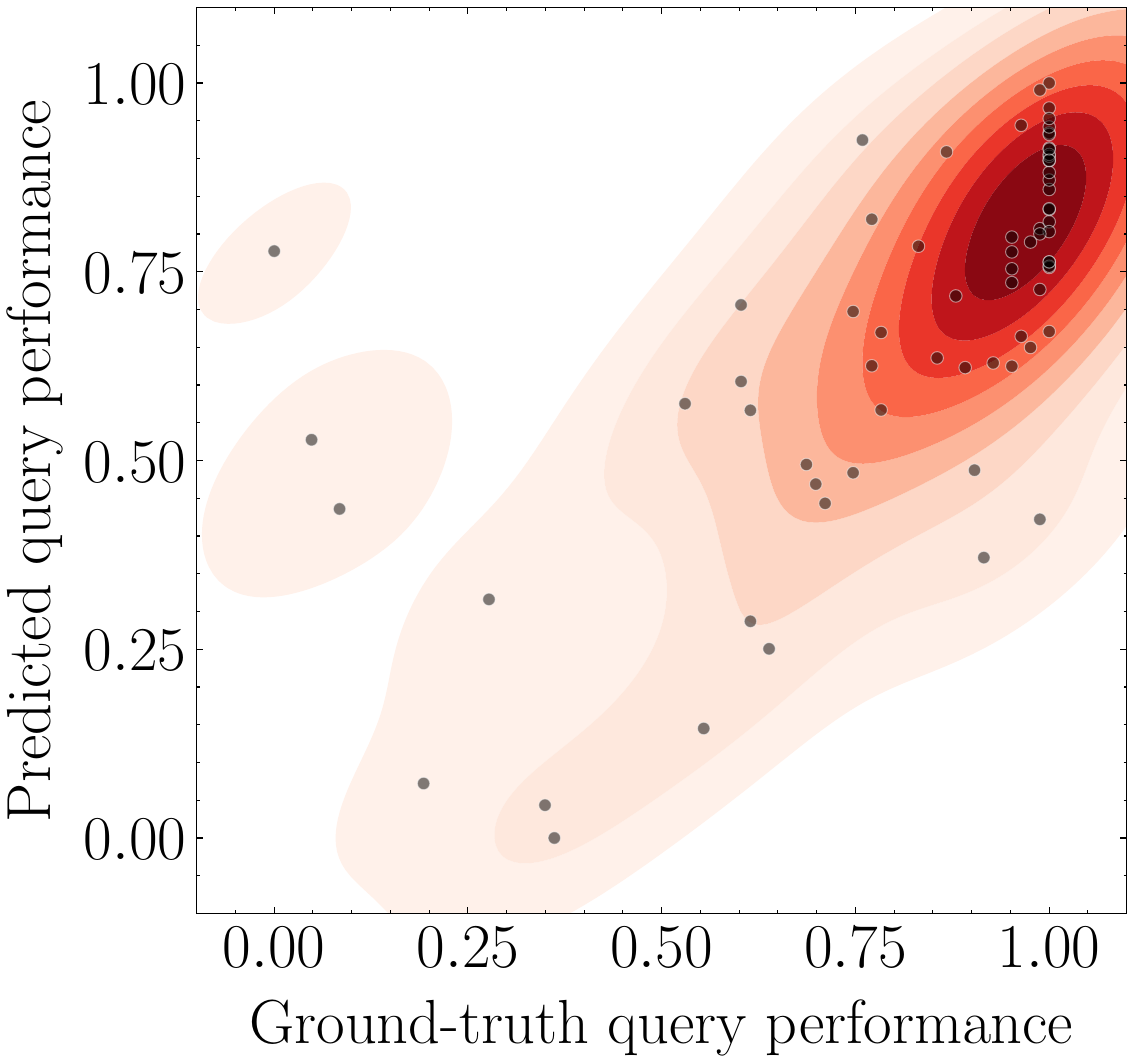}
}
\vspace{-0.3cm}
\caption{Correlation plots between the ground-truth performance given by the P@$\mathbf{100}$ measure for the system of Revaud et al.~\cite{Revaud-ICCV-2019}, and the query performance predicted by the meta-regressor. The intensity of red is proportional to the density of points in the corresponding region.}
\Description{Correlation plots between the ground-truth performance given by the P@$\mathbf{100}$ measure for the system of Revaud et al.~\cite{Revaud-ICCV-2019}, and the query performance predicted by the meta-regressor. The intensity of red is proportional to the density of points in the corresponding region.}
\label{fig_corr_plots}
\vspace{-0.3cm}
\end{figure}

\subsection{Generic discussion}
\label{sec_results_overall} 

Our empirical results reveal that many predictors are only suitable for certain data sets (for example, image difficulty for PASCAL VOC), ground-truth measures (for example, auto-encoders for P@$100$), retrieval systems (for example, score variance for the system of Revaud et al.~\cite{Revaud-ICCV-2019}) or correlation coefficients (for example, the correlation-based CNN for the Pearson correlation). Hence, the results demonstrate the importance of building a comprehensive benchmark based on multiple data sets, retrieval systems and metrics, to establish the generalization capacity of predictors and their robustness to variations of the above components.

The iQPP benchmark includes a high variety of scenarios, being difficult to find a single predictor that is consistently better over all scenarios. However, there are some generic observable trends. First, we notice that post-retrieval predictors generally obtain higher correlations than pre-retrieval predictors. Second, we observe that supervised predictors generally outperform the unsupervised ones. The meta-regressor appears to be the best predictor, being closely followed by the correlation-based CNN, confirming the observed trends. Since high Pearson or Kendall $\tau$ correlation scores can be sometimes misleading, we illustrate the correlation plots between the ground-truth query performance and the performance predicted by the meta-regressor in Figure \ref{fig_corr_plots}. We observe that the plots correspond to the reported numbers, i.e.~the higher correlations on Caltech-101, ROxford5k and RParis6k, and the lower correlation on PASCAL VOC 2012 are visually confirmed by the plots in Figure \ref{fig_corr_plots}.


\section{Conclusion}

In this paper, we introduced the first benchmark for image QPP, comprising four data sets, two retrieval systems and twelve query performance predictors. We studied a wide variety of query performance predictors for CBIR, including state-of-the-art methods \cite{Ionescu-CVPR-2016,Soviany-CVIU-2021,Sun-TIP-2018}, adaptations of text QPP methods \cite{Cummins-SIGIR-2011}, as well as novel proposals. Our benchmark shows that the problem of QPP in image search is still open, as none of the predictors obtained high performance across all data sets and retrieval methods. The empirical results show that our new benchmark exhibits a high variety of evaluation scenarios, representing a real challenge for current and future work on QPP. We thus envision our benchmark as a stepping stone for future research on QPP in the image domain.

In future work, we aim to increase the value of our benchmark by expanding the pool of data sets and retrieval methods. Furthermore, we aim to study novel predictors by leveraging the insights gained from our current experiments. We also plan to delve deeper into the analysis of queries, and separately study easy and hard queries. 

\bibliographystyle{ACM-Reference-Format}
\bibliography{references}


\end{document}